# A generalized mean field algorithm for variational inference in exponential families

**Eric P. Xing**
Computer Science Division
University of California
Berkeley, CA 94720

**Michael I. Jordan**
Computer Science and Statistics
University of California
Berkeley, CA 94720

**Stuart Russell**
Computer Science Division
University of California
Berkeley, CA 94720

## Abstract

We present a class of generalized mean field (GMF) algorithms for approximate inference in exponential family graphical models which is analogous to the generalized belief propagation (GBP) or cluster variational methods. While those methods are based on overlapping clusters, our approach is based on nonoverlapping clusters. Unlike the cluster variational methods, the approach is proved to converge to a globally consistent set of marginals and a lower bound on the likelihood, while providing much of the flexibility associated with cluster variational methods. We present experiments that analyze the effect of different choices of clustering on inference quality, and compare GMF with belief propagation on several canonical models.

## 1 Introduction

The variational approach to probabilistic inference involves converting the inference problem into an optimization problem, by approximating the feasible set or the function to be optimized (or both), and solving the relaxed optimization problem. Thus, given a probability distribution $p(\mathbf{x}|\theta)$ which factors according to a graph, the variational methods yield approximations to marginal probabilities via the solution to an optimization problem that generally exploits some of the graphical structure. The earliest variational inference methods were based on the use of a family of tractable distributions $q(\mathbf{x}|\gamma)$, where $\gamma$ are a set of free "variational parameters." In this case a simple appeal to Jensen's inequality produces a relaxed optimization problem that determines how to set the variational parameters (Jordan et al., 1999). We will refer to such methods as "mean field methods," a terminology that reflects the classical setting in which $q(\mathbf{x}|\gamma)$ is taken to be a completely factorized distribution. In general, the derivation via Jensen's inequality shows that this class of algorithms yields a lower bound on the likelihood.

More recently, Yedidia et al. (2001) realized that Pearl's belief propagation (BP) algorithm—when applied to general loopy graphs—is also a variational algorithm. The inference problem is transformed to an optimization functional—the "Bethe free energy"—that imposes local consistency on the approximate marginals. The resulting marginals do not, however, need to be globally consistent, so that the Jensen inequality argument no longer applies (and thus the approximation does not yield a lower bound to the likelihood and may not converge). An advantage of this approach is the simplicity of the algorithm. Moreover, Yedidia et al. showed how to derive generalized belief propagation (GBP) algorithms, in which the variational relaxation is based on overlapping clusters of variables. The flexibility provided by the ability to choose clusters of varying sizes is a significant important step forward.

Mean field methods can also provide flexibility via the choice of approximating distribution $q(\mathbf{x}|\gamma)$, and so-called "structured mean field methods" have been based on choosing $q(\mathbf{x}|\gamma)$ to be a tree or some other sparse subgraph of the original graph to which an exact inference algorithm such as the junction tree algorithm can be feasibly applied (Saul and Jordan, 1996). Recently, Wiegerinck presented a general framework for structured mean field methods involving arbitrary clusterings (Wiegerinck, 2000). In particular, his approach allows the use of overlapping clusters, which leads to a set of mean field equations reminiscent of a junction tree algorithm. Although there continue to be developments in this area (e.g., El-Hay. and Friedman, 2001, Bishop et al., 2002), it is fair to say that in practice the use of mean-field-based variational methods requires substantial mathematical skill and that a systematic approach with the generality, flexibility and ease of implementation of GBP has yet to emerge. In this paper we describe a Generalized Mean Field method that aims to fill this gap. The approach yields a simple general methodology that applies to a wide range of models. To obtain the desired simplicity our approach makes use of *nonoverlapping* clusters, specializing Wiegerinck's general approach, and yielding a method that is somewhat reminiscent of block methods in MCMC such as Swendsen-Wang (Swendsen and Wang, 1987).

Note that the choice of clusters is generally done manually both within the GBP tradition and the meanfield tradition. Another reason for our interest in nonoverlapping clusters is that it suggests algorithms for automatically choosing clusters based on spectral graph partitioning ideas. Although not the focus of the current paper, we discuss some of the possibilities in Sec. 6.

Given an arbitrary decomposition of the original model into disjoint clusters, the algorithm that we present computes the posterior marginal for each cluster given its own evidence and the *expected sufficient statistics*, obtained from its neighboring clusters, of the variables in the cluster's Markov blanket. The algorithm operates in an iterative, message-passing style until a fixed point is reached. We show that under very general conditions on the nature of the inter-cluster dependencies, the cluster marginals retain exactly the intra-cluster dependencies of the original model, which means that the inference problem within each cluster can be solved independently of the other clusters (given the Markov blanket messages) by any inference method.

One way to understand the algorithm is to consider a situation in which all the Markov blanket variables of each cluster are observed. In that case, the joint posterior decomposes:

$$p(\mathbf{x}_{C_1},\ldots,\mathbf{x}_{C_n}|\mathbf{x}_E) = \prod_i p(\mathbf{x}_{C_i}|MB(\mathbf{x}_{C_i})),$$

where $MB(\mathbf{x}_{C_i})$ denotes the Markov blanket of cluster $C_i$. GMF approximates this situation, using the expected Markov blanket (obtained from neighboring clusters) instead of an observed Markov blanket and iterating this process to obtain the best possible "self-consistent" approximation.

In its use of expectations in messages between clusters, GMF resembles the expectation propagation (EP) algorithm (Minka, 2001), but in the basic algorithm EP's messages convey the influence of only a single variable. In providing a generic variational algorithm that can be applied to a broad range of models with convergence guarantees, GMF resembles VIBES (Bishop et al., 2002), but VIBES is based on a decomposition into individual variables whereas GMF allows arbitrary disjoint sets. Thus GMF is a generic algorithm suitable for approximate inference in large, complex probability models.

## 2 Notation and background

We consider a graph (directed or undirected) $G = (V, L)$, where $V$ denotes the set of nodes (vertices) and $L$ the set of edges (links) of the graph. Let $X_n$ denote the random variable associated with node $n$, for $n \in V$, let $\mathbf{X}_C$ denote the subset of variables associated with a subset of nodes $C$, for $C \subseteq V$, and let $\mathbf{X} = \mathbf{X}_V$ denote the collection of all variables associated let with the graph. We refer to a graph $H = (V, L')$, where $L' \subset L$, as a *subgraph* of $G$. We use $\mathcal{C} = \{C_1, C_2, \ldots, C_I\}$ to denote a disjoint partition (or, a *clustering*) of all nodes in graph $G$, where $C_i$ refers to the set of indices of nodes in cluster $i$; likewise, $\mathcal{D} = \{D_1, D_2, \ldots, D_K\}$ denotes a set of *cliques* of $G$. For a given clustering, we define the *border clique set* $\mathcal{B}_i$ as the set of cliques that intersect with but are not contained in cluster $i$; and the *neighbor cluster set* $\mathcal{N}_i$ as the set of clusters that contain nodes connected to nodes in cluster $i$. For undirected graphs, the *Markov blanket* of a cluster $i$ ($MB_i$) is the set of all nodes outside $C_i$ that connect to some node in $C_i$, and, for directed graphs, the Markov blanket is the set of all nodes that are parents, children, or co-parents of some node in $C_i$ (Fig. 1). Clusters that intersect with $MB_i$ are called the *Markov blanket clusters* ($MBC_i$) of $C_i$.

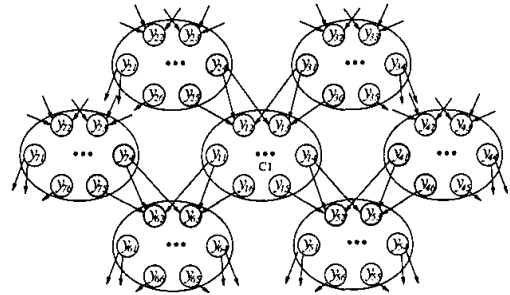

Figure 1: The Markov blanket $MB_1$ (blue-shaded nodes) of cluster 1 in a directed graph. Shaded blobs constitute $MBC_1$.

### 2.1 Exponential representations

For undirected graphical models, the family of joint probability distributions associated with a given graph can be parameterized in terms of a set of *potential functions* associated with a set of cliques in the graph. For a set of cliques $\mathcal{D}$ associated with an undirected graph, let $\phi = \{\phi_\alpha | \alpha \in \mathcal{D}\}$ denote the set of potential functions defined on the cliques, and $\boldsymbol{\theta} = \{\theta_\alpha | \alpha \in \mathcal{D}\}$ the set of parameters associated with these potential functions (for simplicity, we label $\phi$ and $\theta$ with the corresponding *clique index*, e.g., $\alpha$, rather than with the clique $D_\alpha$ itself). The family of joint distributions determined by $\phi$ can be expressed as follows:

$$p(\mathbf{x}|\boldsymbol{\theta}) = \exp\{\sum_\alpha \theta_\alpha \phi_\alpha(\mathbf{x}_{D_\alpha}) - A(\boldsymbol{\theta})\} \qquad (1)$$

where $A(\boldsymbol{\theta})$ is the *log partition function*. We also define the *energy*, $E(\mathbf{x}) = -\sum_\alpha \theta_\alpha \phi_\alpha(\mathbf{x}_{D_\alpha})$, for state $\mathbf{x}$.

For directed graphical models, in which the joint probability is defined as $p(\mathbf{x}) = \prod_i p(x_i|\mathbf{x}_{\pi_i})$, we transform the underlying directed graph into a *moral graph*, and set the potential functions equal to the negative logarithm of the local conditional probabilities $p(x_i|\mathbf{x}_{\pi_i})$. In the sequel, we will focus on models based on *conditional exponential families*. That is, the conditional distributions $p(x_i|\mathbf{x}_{\pi_i})$ can be expressed as:

$$p(x_i|\mathbf{x}_{\pi_i}) = u(x_i)\exp\{\theta_i^T \phi_i(x_i,\mathbf{x}_{\pi_i}) - A(\theta_i)\}, \quad (2)$$

where $\phi_i(x_i,\mathbf{x}_{\pi_i})$ is a vector of potentials associated with variable set $\{x_i,\mathbf{x}_{\pi_i}\}$.

## 2.2 Cluster-factorizable potentials

Given a clustering $\mathcal{C}$, some cliques in $\mathcal{D}$ may intersect with multiple clusters (Fig. 2). *Cluster-factorizable potentials* are potential functions which take the form $\phi_\beta(\mathbf{x}_{D_\beta}) = F_\beta(\phi_{\beta_i}(\mathbf{x}_{D_\beta \cap C_i}), \ldots, \phi_{\beta_j}(\mathbf{x}_{D_\beta \cap C_j}))$, where $F(\cdot)$ is a (multiplicatively, or additively) factorizable function over its arguments; i.e., in the case of two clusters, $F(a,b) = a \times b$ or $a+b$. Factorizable potentials are common in many model classes. For example, the classical Ising model is based on singleton and pairwise potentials of the following factorizable form: $\phi(x_i) = \theta_i x_i$, $\phi(x_i,x_j) = \theta_{ij}x_ix_j$; higher-order Ising models and general discrete models also admit factorizable potentials; conjugate exponential pairs, such as the Dirichlet-multinomial, linear-Gaussian, etc., are also factorizable; finally, for logistic functions and other generalized linear models (GLIMs) that are not directly factorizable, it is often possible to obtain a factorizable variational transformation in the exponential family that lower bounds the original function (Jaakkola and Jordan, 2000); otherwise (e.g., tabular potentials over a clustering of variables), we may overcome this problem by avoiding picking a clustering in which these potentials are on the cluster boundaries. We will see that cluster-factorizable potentials allow the decoupling of the computation of expected potentials.

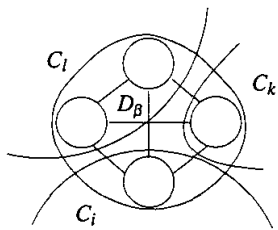

Figure 2: A clique $D_\beta$ intersecting with three clusters $\{C_i, C_j, C_k\}$ in an undirected graph.

## 3 Mean Field Approximation

Recall that the mean field approximation refers to a class of variational approximation methods that approximate the true distribution $p(\mathbf{x}|\theta)$ on a graph $G$ with a simpler distribution, $q(\mathbf{x}|\gamma)$, for which it is feasible to do exact inference. We call the families of such distributions *tractable families*. A tractable family usually corresponds to a subgraph $H$ of $G$.

### 3.1 Naive mean field approximation

The naive mean field approximation makes use of a subgraph that is completely disconnected. Thus, the approximating distribution is fully factorized:

$$q(\mathbf{x}) = \prod_{i \in V} q_i(x_i). \quad (3)$$

For example, to use this family of distributions to approximate the joint probability of the Boltzmann machine: $p(\mathbf{x}) = \frac{1}{Z}\exp\{\sum_{i<j}\theta_{ij}x_ix_j + \sum_i \theta_{i0}x_i\}$, one defines $q_i(x_i) = \mu_i^{x_i}(1-\mu_i)^{1-x_i}$, where the $\mu_i$ are the variational parameters). Minimizing the Kullback-Leibler (KL) divergence between $q$ and $p$ one obtains the classical "mean field equations":

$$\mu_i = \sigma\Big(\sum_{j \in \mathcal{N}_i} \theta_{ij}\mu_j + \theta_{i0}\Big), \quad (4)$$

where $\sigma(z) = 1/(1+e^{-z})$ is the logistic function, and $\mathcal{N}_i$ is the set of nodes neighboring $i$.

### 3.2 Generalized mean field theory

Given a (disjoint) clustering $\mathcal{C}$, we define a cluster-factorized distribution as $q(\mathbf{x}) = \prod_{C_i \in \mathcal{C}} q_i(\mathbf{x}_{C_i})$, where $q_i(\mathbf{x}_{C_i}) = \exp\{-E_i'(\mathbf{x}_{C_i})\}, \forall C_i \in \mathcal{C}$, are free distributions to be optimized. As discussed in the appendix, this optimization problem can be cast as that of maximizing a lower bound of the likelihood with respect to all valid cluster marginals respecting a given clustering $\mathcal{C}$. The solution to this problem leads to the generalized mean field theorem that we present in this section.

To make the exposition of the theorem and the resulting algorithm simple, we introduce some definitions.

**Definition 1.** (Mean field factor): For a factorizable potential $\phi_\beta(\mathbf{x}_{D_\beta})$, let $I_\beta$ denote the set of indices of those clusters that have nonempty intersection with $D_\beta$. Thus, $\phi_\beta(\mathbf{x}_{D_\beta})$ has as factors the potentials $\phi_{\beta_i}(\mathbf{x}_{C_i \cap D_\beta}), \forall i \in I_\beta$. Then, the *mean field factor* $f_{i\beta}$ is defined as:

$$f_{i\beta} \triangleq f_{i\beta}(\mathbf{x}_{C_i \cap D_\beta}) \triangleq \langle \phi_{\beta_i}(\mathbf{x}_{C_i \cap D_\beta})\rangle_{q_i}, \quad \text{for } i \in I_\beta \quad (5)$$

where $\langle \cdot \rangle_{q_i}$ denotes the expectation with respect to $q_i$.

**Definition 2.** (Generalized mean fields): For any cluster $C_j$ in a given variable partition, the set of mean field factors associated with the nodes in its *Markov blanket* is referred as the *generalized mean fields* of cluster $C_j$:

$$\mathcal{F}_j \triangleq \{f_{i\beta} : D_\beta \in \mathcal{B}_j, i \in I_\beta, i \neq j\}. \quad (6)$$

Now we are ready to state the following GMF theorem, the proof of which is provided in the Appendix.

**Theorem 3.** *For a general undirected probability model $p(\mathbf{x}_H, \mathbf{x}_E)$ where $\mathbf{x}_H$ denotes hidden nodes and $\mathbf{x}_E$ denotes evidence nodes, and a clustering $\mathcal{C} : \{\mathbf{x}_{H,C_i}, \mathbf{x}_{E,C_i}\}_{i=1}^I$ of both hidden and evidence nodes, if all the potential functions that cross cluster borders are cluster-factorizable, then the generalized mean field approximation to the joint posterior $p(\mathbf{x}_H|\mathbf{x}_E)$ with respect to clustering $\mathcal{C}$ is a product of cluster marginals $q^{GMF}(\mathbf{x}_H) = \prod_{C_i \in \mathcal{C}} q_i^{GMF}(\mathbf{x}_{H,C_i})$ satisfying the following generalized mean field equations:*

$$q_i^{GMF}(\mathbf{x}_{H,C_i}) = p(\mathbf{x}_{H,C_i}|\mathbf{x}_{E,C_i}, \mathcal{F}_i), \quad \forall i. \quad (7)$$

**Remark 1.** Note that each variational cluster marginal is isomorphic to the isolated model fragment corresponding to original cluster posterior given the intra-cluster evidence and the *generalized mean fields* from outside the cluster. Thus, each variational cluster marginal inherits all local dependency structures inside the cluster from the original model.

The mean field equations in Theorem 3 are analogous to naive mean field approximation. The *generalized mean fields* appearing in Eq. (7) play a role that is similar to the conventional mean field, now applying to the entire cluster rather than a single node, and conducting probabilistic influence from the remaining part of the model to the cluster. It is easy to verify that when the clusters reduce to singletons, Eq. (7) is equivalent to the classical mean field equation Eq. (4). From a conditional independence point of view, the generalized mean fields can be also understood as an *expected Markov blanket* of the corresponding cluster, rendering its interior nodes conditionally independent of the remainder of the model and hence localizing the inference within each cluster given its generalized mean fields.

Mean field approximation for directed models is also covered by Theorem 3. This is true because any directed network can be converted into an undirected network via moralization, and designation of the potentials as local conditional probabilities. The following corollary make this generalization explicit:

**Corollary 4.** *For a directed probability model $p(\mathbf{x}_H, \mathbf{x}_E) = \prod_i p(x_i|\mathbf{x}_{\pi_i})$ and a given disjoint variable partition, if all the local conditional models $p(x_i|\mathbf{x}_{\pi_i})$ across the cluster borders admit cluster-factorizable potentials, then the generalized mean field approximation to the original distribution has the following form: $q^{GMF}(\mathbf{x}_H) = \prod_{C_i \in \mathcal{C}} q_i^{GMF}(\mathbf{x}_{H,C_i})$, and*

$$q_i^{GMF}(\mathbf{x}_{H,C_i}) = p(\mathbf{x}_{H,C_i}|\mathbf{x}_{E,C_i}, \mathcal{F}_i), \quad \forall i, \quad (8)$$

*where $\mathcal{F}_i$ refers to the generalized mean fields of the exterior parents, children and co-parents of the variables in cluster $i$.*

These theorems make it straightforward to obtain generalized mean field equations. All that is needed is to decide on a subgraph and a variable clustering, to identify the Markov blanket of each cluster, and to plug in the mean fields of the Markov blanket variables according to Eqs. (7) or (8). We illustrate the application of the generalized mean field theorem to several typical cases—undirected models, directed models, and models that combine continuous and discrete random variables.

**Example 1.** *(2-d nearest-neighbor Ising model):* For a 2-d nearest neighbor Ising model, we can pick a subgraph whose connected components are square blocks of nodes in the original graph (Fig. 3). The cluster marginal of a square block $G_k$ is simply $q(\mathbf{x}_{G_k}) = \exp\{\sum_{(ij) \in L(G_k)} \theta_{ij} x_i x_j + \sum_{i \in V(G_k)} \theta_{i0} x_i + \sum_{(ij) \in L(G), j \in MB(G_k)} \theta_{ij} \langle x_j \rangle x_i\}$, an Ising model of smaller size, with singleton potentials for the peripheral nodes adjusted by the mean fields of the adjacent nodes outside the block (which are the MB of $\mathbf{x}_{G_k}$). ◇

**Example 2.** *(factorial hidden Markov models):* For the fHMM, whose underlying graph consists of multiple chains of discrete hidden Markov variables coupled by a sequence of output nodes, taken to be linear-Gaussian for concreteness, a possible subgraph that defines a tractable family is shown in Figure 5, in which we retain only the edges within each chain of the original graph. Given a clustering $\mathcal{C}$, in which each cluster $k$ contains a subset of HMM chains $c_k$ (the dashed boxes in Fig. 5), the MB of each cluster consists of all nodes outside the cluster. Hence the cluster marginal of $c_k$ is: $q(\{\mathbf{x}^{(m_i)}\}_{i \in c_k}) \propto \prod_{i \in c_k} p(\mathbf{x}^{(m_i)}) p(\mathbf{y}|\{\mathbf{x}^{(m_i)}\}_{i \in c_k}, \{f(\mathbf{x}^{(m_j)})\}_{j \in c_l, l \neq k})$, where $\mathbf{x}^{(m_i)}$ denotes variables of chain $m_i$, $p(\mathbf{x}^{(m_i)})$ is the usual HMM of a single chain, and $p(\mathbf{y}|\cdot)$ is linear-Gaussian. When each $c_k$ contains only a single chain, we recover the structured variational inference equations in Ghahramani and Jordan (1997). ◇

**Example 3.** *(Variational Bayesian learning):* Following the standard setup in Ghahramani and Beal (2000), we have a *complete data likelihood* $P(\mathbf{x}, \mathbf{y}|\theta)$, where $\mathbf{x}$ is hidden, and a *prior* $p(\theta|\eta, \nu)$, where $\eta, \nu$ are *hyperparameters*. Partitioning all domain variables into two clusters, $\{\mathbf{x}, \mathbf{y}\}$ and $\{\theta\}$, if the potential function at the cluster border, $\phi(\mathbf{x}, \theta)$ is factorizable (which is equivalent to the condition of *conjugate exponentiality* in Ghahramani and Beal), we obtain the following cluster marginals using Corollary 4:

$$q(\theta) = p(\theta|\eta, \nu, f(\mathbf{x}), \mathbf{y}) \propto p(f(\mathbf{x}), \mathbf{y}|\theta) p(\theta|\eta, \nu)$$
$$q(\mathbf{x}) = p(\mathbf{x}|\mathbf{y}, f(\theta)).$$

These coupled updates are identical to the variational Bayesian learning updates of Ghahramani and Beal. ◇

## 4 A generalized mean field algorithm

Eqs. (7) and (8) are a coupled set of nonlinear equations, which are solved numerically via asynchronous iteration until a fixed point is reached. This iteration constitutes a simple, message-passing style, Generalized Mean Field algorithm.

---

**GMF** ( model: $p(\mathbf{x}_H, \mathbf{x}_E)$, partition: $\{\mathbf{x}_{H,C_i}, \mathbf{x}_{E,C_i}\}_{i=1}^I$)
**Initialization**
- Randomly initialize the hidden nodes at the border of cluster $i$, $\forall i$.
- Initialize $f_{i,\beta}^0$ by evaluating the potentials using the current values of the associated nodes.
- Initialize $\mathcal{F}_i^0$ with the current $f_{i,\beta}^0$.

**While** not converged
　**For** $i = 1 : I$
　- Update $q_i^{t+1}(\mathbf{x}_{H,C_i}) = p(\mathbf{x}_{H,C_i}|\mathbf{x}_{E,C_i}, \mathcal{F}_i^t)$.
　- Compute the mean field factors $f_{i,\beta}^{t+1}$ of all potential factors at the border of $C_i$ via local inference using $q_i^{t+1}$ as in Eq. (5).
　- Send the $f_{i,\beta}^{t+1}$ messages to all Markov blanket clusters of $i$ by updating the appropriate elements in their GMFs: $\mathcal{F}_j^t \to \mathcal{F}_j^{t+1}, \forall j \in MBC_i$.
　**End**
**Return** $q(\mathbf{x}_H) = \prod_i q_i(\mathbf{x}_{H,C_i})$, the GMF approximation

---

**Remark 2.** Note that the r.h.s. of Eqs. (7) and (8) do not depend on $q_i$, thus the update is a form of coordinate ascent in the factored model space (i.e., we fix all $q_j(\mathbf{x}_{H,C_j}), j \neq i$ and maximize with respect to $q_i(\mathbf{x}_{H,C_i})$ at each step). Indeed, we have the following convergence theorem.

**Theorem 5.** *The GMF algorithm is guaranteed to converge to a local minimum, which is a lower bound for the likelihood of the model.*

Theorem 5 is an important consequence of the use of a *disjoint* variable partition underlying the variational approximate distribution. It distinguishes GMF from other variational methods such as GBP (Yedidia et al., 2001), or the general case in Wiegerinck's framework (Wiegerinck, 2000), in which overlapping variable partitions are used, and which optimize an approximate free energy function with respect to marginals which must satisfy local constraints.

The complexity of each iteration of GMF is exponential in the tree-width of the *local* networks of each cluster of variables, since inference is reduced to local operations within each cluster.

Since GMF is guaranteed to converge to a local optimum, in practice it can be performed in a stochastic multiple-initialization setting similar to the usual practice in EM, to increase the chance of finding a better local optimum.

## 5 Experimental results

Although GMF supports several types of applications, such as finding bounds on the likelihood or log-partition function, computation of approximate marginal probabilities, and parameter estimation, in this paper we focus solely on the quality of approximate marginals. We have performed experiments on three canonical models: a nearest neighbor Ising model (IM), a sigmoid network (SN), and a factorial HMM (fHMM); and we have compared performance of GMF using different tractable families (specifically, using variable clusterings of different granularity) with regard to the accuracy on single-node marginals. To assess the error, we use an $L_1$-based measure

$$\frac{1}{\sum_{i=1}^N M_i} \sum_{i=1}^N \sum_{k=1}^{M_i} |p(x_i = k) - q(x_i = k)|,$$

where $N$ is the total number of variables, and $M_i$ is the number of (discrete) states of the variable $x_i$. The exact marginals are obtained via the junction tree algorithm. We also compare the performance with the belief propagation (BP) algorithm, especially in cases where BP is expensive, and examine whether GMF provides a reasonably efficient alternative.

We use randomly generated problems for IM and SN and real data for fHMM. For the first two cases, in any given trial we specify the distribution $p(\mathbf{x}|\theta)$ by a random choice of the model parameter $\theta$ from a uniform distribution. For models with observable output (i.e., evidence), observations were sampled from the random model. Details of the sampling are specified in the tables presenting the results. For each problem, 50 trials were performed. The fHMM experiment was performed on models learned from a training data set.

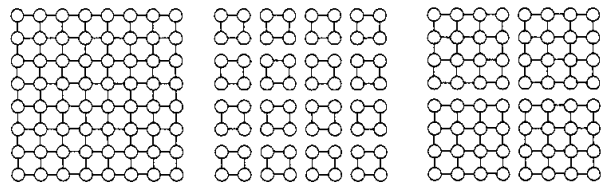

Figure 3: Ising model and GMF approximations.

**Ising models:** We used an 8 × 8 grid with binary nodes. Two different tractable models were used for the GMF approximation, one based on a clustering of 2 × 2 blocks, the other 4 × 4 blocks (Fig. 3). The results on strongly attractive and repulsive Ising models (which are known to be difficult for naive MF) are reported in Table 1. The rightmost column also shows the mean CPU time (in seconds).

Table 1: $L_1$ errors on nearest neighbor Ising models.
Upper panel: attractive IM ($\theta_{i0} \in (-0.25, 0.25), \theta_{ij} \in (0, 2)$);
Lower panel: repulsive IM ($\theta_{i0} \in (-0.25, 0.25), \theta_{ij} \in (-2, 0)$).

| Algorithm | Mean ± std | Median | Range | time |
|---|---|---|---|---|
| 2 × 2 GMF | 0.366±0.054 | 0.382 | [0.276,0.463] | 2.0 |
| 4 × 4 GMF | 0.193±0.103 | 0.226 | [0.004,0.400] | 29.4 |
| BP | 0.618±0.304 | 0.663 | [0.054,0.995] | 17.9 |
| GBP | 0.003±0.002 | 0.002 | [0.000,0.005] | 166.3 |
| 2 × 2 GMF | 0.367±0.052 | 0.383 | [0.279,0.449] | 1.2 |
| 4 × 4 GMF | 0.185±0.102 | 0.161 | [0.009,0.418] | 22.1 |
| BP | 0.351±0.286 | 0.258 | [0.009,0.954] | 14.3 |
| GBP | 0.003±0.003 | 0.003 | [0.000,0.014] | 117.5 |

As expected, GMF using a clustering with fewer nodes decoupled yields more accurate estimates than a clustering in which more nodes are decoupled, albeit with increased computational complexity. Overall, the performance of GMF is better than that of BP, especially for the attractive Ising model. For this particular problem, we also compared to the GBP algorithm, which also defines beliefs on larger subsets of nodes, with a more elaborate message-passing scheme. We found that for Ising models, GBP performs significantly better than the other methods, but at a cost of significantly longer time to convergence.

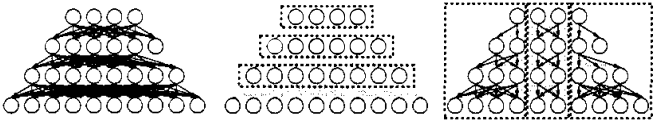

Figure 4: Sigmoid network and GMF approximations.

**Sigmoid belief networks:** The two sigmoid networks we studied are comprised of three hidden layers (18 nodes), with or without a fourth observed layer (10 nodes), respectively. We used a row clustering and a block clustering of nodes as depicted in Figure 4 for GMF. Table 2 summarizes the results.

Table 2: $L_1$ errors on sigmoid networks ($\theta_{ij} \in (0,1)$).
Upper: hidden layers only; Lower: with observation layer.

| Algorithm | Mean ± std | Median | Range | time |
|---|---|---|---|---|
| block GMF | 0.013±0.004 | 0.013 | [0.006,0.032] | 6.8 |
| row GMF | 0.172±0.036 | 0.175 | [0.100,0.244] | 0.5 |
| BP | 0.273±0.025 | 0.271 | [0.227,0.346] | 9.2 |
| block GMF | 0.018±0.009 | 0.014 | [0.009,0.038] | 8.4 |
| row GMF | 0.061±0.021 | 0.059 | [0.023,0.145] | 0.7 |
| BP | 0.187±0.044 | 0.189 | [0.096,0.312] | 139.2 |

For the network without observations, the block GMF, which retains a significant number of edges from the original graph, is more accurate by an order of magnitude than the row GMF, which decouples the original network completely. Interestingly, when a bottom layer of observed nodes is included in the network, a significant improvement of approximation accuracy is seen for the row GMF, but it still does not surpass the block GMF. The performance of BP is poor on both problems, and the time complexity scales up significantly for the network with the observation layer, because of the large fan-in associated with the nodes in the bottom layer.

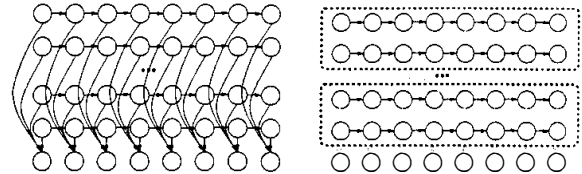

Figure 5: An fHMM and a GMF approximation (illustrative graph; the actual model contains 6 chains and 40 steps).

**Factorial HMM:** We studied a 6-chain fHMM, with (6-dimensional) linear-Gaussian emissions, ternary hidden state and 40 time steps. The model was trained using the EM algorithm (with exact inference) on 40 Bach Chorales from the UCI Repository. Inference was performed with the trained model on another 18 test Chorales. GMF approximations were based on clusterings in which each cluster contains either singletons (i.e., naive mean field), one hidden Markov chain, two chains, or three chains, respectively. The statistics of the $L_1$ errors are presented in Table 3.

Table 3: $L_1$ errors on factorial HMM

| Algorithm | Mean ± std | Median | Range | time |
|---|---|---|---|---|
| naive MF | 0.254±0.095 | 0.269 | [0.083,0.397] | 9.8 |
| 1-chain GMF | 0.237±0.107 | 0.233 | [0.029,0.392] | 14.3 |
| 2-chain GMF | 0.092±0.081 | 0.064 | [0.019,0.314] | 5.6 |
| 3-chain GMF | 0.118±0.092 | 0.089 | [0.035,0.357] | 15.6 |
| BP | 0 | 0 | - | 106.2 |

Since the moral graph of a fHMM is a clique tree, BP is exact in this case, but the computational complexity grows exponentially with the number of chains and the cardinality of the variables, hence BP cannot scale to large models. Using GMF, we obtain reasonable accuracy, which in general increases with the granularity of the variable clustering. The 2-chain GMF appears to be a particularly good granularity of clustering in this case, leading to both better estimation and faster convergence.

In summary, GMF shows reasonable performance in all three of the canonical models we tested, and provides a flexible way to trade off accuracy for computation time. It is guaranteed to converge, and the computational complexity is determined by the treewidth of the subgraph. BP, on the other hand, may fail to converge. Furthermore, the complexity of computing the message is exponential in the size of the maximal clique in the moralized graph, which makes it very expensive in directed models with dense local dependencies.

## 6 Choice of clusters

One reason for our focus on disjoint partitions has been the simplicity and ease-of-implementation of the resulting algorithm. But it is also the case that the use of disjoint partitions opens up an interesting new set of research problems involving the choice of clusters. Intuition suggests that one possible definition of a good partitions is one in which many edges are cut, with relatively small parameter values across the cut. In this setting we would expect to have concentration of the expectations of the potentials—the "mean fields" would be well determined.

In Xing and Jordan (2003) we explore this idea by combining the GMF algorithm with combinatorial optimization methods for graph partition. We have found that, depending on the connectivity and coupling strength of the graphical model, various automatic graph partition schemes can yield effective clusterings. For example, for densely connected graph with weak coupling, a max-cut indeed leads to improved approximation of marginal probabilities when compared to naive mean field and other simple fixed partition schemes. On the other hand, for a graph with relatively sparse connectivity, and strong coupling, a min-cut of the graph leads to better estimation of marginals, possibly due to an improved ability to capture the dependency structure within each cluster, in a manner analogous to the cut-set conditioning methods used for exact inference. These promising results open up the possibility for a fully autonomous variational inference algorithm for complex models based on automatic node partition of a graphical model and GMF approximation as illustrated in the following flowchart in Figure 6. A prototype implementation of such an algorithm is available at: http://www.cs.berkeley.edu/~epxing/GMF.zip.

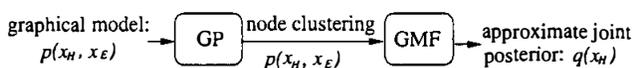

Figure 6: Flowchart of a autonomous variational inference algorithm.

## 7 Discussion

We have presented a generalized mean field approach to probabilistic inference in graphical models, in which a complex probability distribution is approximated via a distribution that factorizes over a disjoint partition of the graph. Locally optimal variational approximations are obtained via an algorithm that performs co-ordinate ascent in a lower bound of the log-likelihood, with guaranteed convergence. For a broad family of models in practical use, we showed that the GMF approximations of the cluster marginals are isomorphic to the original model in the sense that they inherit all of its intra-cluster dependencies. Moreover, these marginals are independent of the rest of the model given the expected potential factors (mean fields) of the Markov blanket of the cluster. The explicit and generic formulation of the "mean fields" in terms of the Markov blanket of variable clusters also leads to a simple, generic, message-passing algorithm for complex models.

Disjoint clusterings have also been used in sampling algorithms to improve mixing rates for large problems. For example, the Swendsen-Wang algorithm (Swendsen and Wang, 1987) samples Ising (or Potts) model at critical temperatures by grouping neighboring nodes with the same spin value, thereby forming random clusters (of coupled spins) that are effectively independent of each other, allowing an MCMC process to collectively sample the spin of each cluster independently and at random. This method often dramatically speeds up the mixing of the MCMC chain. Gilks et al. (1996) also noted that when variables are highly correlated in the stationary distribution, blocking highly correlated components into higher-dimensional components may improve mixing. However, in the sampling framework, clustering are usually obtained dynamically, based on the coupling strength, rather than the topology of the network.

There are a number of possible extensions of the research reported here. First, it is of interest to develop automatic methods for choosing clusters in variational approximations. As we have already discussed, spectral graph partitioning can be adapted for this purpose in the case of GMF methods. It is also possible to make use of the framework of probabilistic relational models and motivate partitions of the random variables using modularities deriving from the model semantics (e.g., class membership). Preliminary results in applying this to a large-scale bioinformatics problem showed that it leads to significantly improvements in performance.

Another possible extension involves the use of higher-order expansions in the basic variational bounds. Leisink and Kappen (2000) have shown how to upgrade first-order variational bounds such as that shown in Eq. (10) to yield higher-order bounds. In particular, the following third-order lower bound can be obtained for the likelihood:

$$p(\mathbf{x}_E) \geq \int d\mathbf{x} \exp\{-E'(\mathbf{x}_H)\}\left[1 - \Delta + \frac{1}{2}\exp(\xi)\Delta^2\right],$$

where $\xi = \frac{1}{3}\langle\Delta^3\rangle/\langle\Delta^2\rangle$, $\Delta = E(\mathbf{x}_H, \mathbf{x}_E) - E'(\mathbf{x}_H)$, and $\langle\cdot\rangle$ denotes expectation over the approximate distribution $q(\mathbf{x}_H) = \exp\{-E'(\mathbf{x}_H)\}$. The optimizer of this lower bound cannot be found analytically. However, we can compute the gradient of the lower bound with respect to $E'_i$ (assuming a cluster-factorized approximate distribution), which requires computation of up to third-order cumulants of the nodes in the bordering cliques in the subgraph. Leisink and Kappen (2000) reported an application of such a strategy to the 2-D lattice model and sigmoid belief network, approximated by a completely disconnected subgraph, and reported significantly improved bounds. In the GMF setting, which uses an approximating subgraph with more structure, the computation of the gradient is even simpler because fewer nodes are involved in the cumulant calculation.

## Acknowledgments

We thank Yair Weiss and colleagues for their generosity in sharing their code for exact inference and GBP on grids. This project was supported by ARO MURI DAA19-02-1-0383 and NSF grant IIS-9988642.

## References

C. Bishop, D. Spiegelhalter, and J. Winn. VIBES: A variational inference engine for Bayesian networks. In *NIPS*, 2002.

T. El-Hay. and N. Friedman. Incorporating expressive graphical models in variational approximations: Chain-graphs and hidden variables. In *UAI*, 2001.

Z. Ghahramani and M. Beal. Propagation algorithms for variational Bayesian learning. In *NIPS*, 2000.

Z. Ghahramani and M. I. Jordan. Factorial hidden markov models. *Machine Learning*, 29:245–273, 1997.

W. R. Gilks, S. Richardson, and D. J. Spiegelhalter, editors. *Markov Chain Monte Carlo in Practice*. Chapman and Hall, 1996.

T. S. Jaakkola and M. I. Jordan. Bayesian logistic regression: A variational approach. *Statistics and Computing*, 10:25–37, 2000.

M. I. Jordan, Z. Ghahramani, T. S. Jaakkola, and L. K. Saul. An introduction to variational methods for graphical models. In M. I. Jordan, editor, *Learning in Graphical Models*. MIT Press, Cambridge, 1999.

M. A. R. Leisink and H. J. Kappen. A tighter bound for graphical models. In *NIPS*, 2000.

T. Minka. Expectation propagation for approximate Bayesian inference. In *UAI*, 2001.

L. K. Saul and M. I. Jordan. Exploiting tractable substructures in intractable networks. In *NIPS*, 1996.

R. Swendsen and J-S Wang. Non-universal critical dynamics in Monte Carlo simulation. *Physical Reviews Letters*, 58:86–88, 1987.

W. Wiegerinck. Variational approximations between mean field theory and the junction tree algorithm. In *UAI*, 2000.

E. P. Xing and M. I. Jordan. Graph partition strategies for generalized mean field inference. In preparation, 2003.

J. S. Yedidia, W. T. Freeman, and Y. Weiss. Understanding belief propagation and its generalizations. In *Distinguished Lecture track, IJCAI*, 2001.

## A  Proof of the GMF theorem

To cast GMF approximation as an optimization problem, we begin with the follow lemma.

**Lemma 6.** *For an arbitrary marginal distribution* $q(\mathbf{x}_H) = \exp\{-E'(\mathbf{x}_H)\}$, *we have the following lower bound:*

$$p(\mathbf{x}_E) \geq \int d\mathbf{x}_H \exp\{-E'(\mathbf{x}_H)\}$$
$$\left(1 - A(\mathbf{x}_E) - (E(\mathbf{x}_H, \mathbf{x}_E) - E'(\mathbf{x}_H))\right), \quad (9)$$

*where* $\mathbf{x}_E$ *denotes observed variables (evidence)* [1].

**Proof.** Using conjugate duality, we have:

$$\exp(x) \geq \exp(\mu)(1 + x - \mu), \quad \forall x, \mu. \quad (10)$$

For a joint distribution $p(\mathbf{x}_H, \mathbf{x}_E) = \exp\{-E(\mathbf{x}_H, \mathbf{x}_E) - A(\mathbf{x}_E)\}$ (where $A(\mathbf{x}_E)$ is the original log-partition function plus the constant evidence potentials), we replace $x$ in Eq. (10) with $-(E(\mathbf{x}_H, \mathbf{x}_E) + A(\mathbf{x}_E))$ and lower bound the joint distribution $p(\mathbf{x}_H, \mathbf{x}_E)$ as follows:

$$p(\mathbf{x}_H, \mathbf{x}_E) \geq q(\mathbf{x}_H)\big(1 - A(\mathbf{x}_E) - (E(\mathbf{x}_H, \mathbf{x}_E) - E'(\mathbf{x}_H))\big),$$

where $E'(\mathbf{x}_H)$ defines a *variational marginal distribution*. Integrating over $\mathbf{x}_H$ on both sides, we obtain the first-order lower bound in Eq. (9). ∎

Given this lower bound, the optimal approximating (GMF) distribution is specified as the solution of the following constrained optimization problem:

$$\{E'^{GMF}_i(\mathbf{x}_{C_i})\}_{C_i \in \mathcal{C}} = \arg\max_{E'_i \in \mathcal{E}(\mathbf{x}_{C_i})}$$
$$\int d\mathbf{x} \exp\left\{-\sum_{C_i \in \mathcal{C}} E'_i(\mathbf{x}_{C_i})\right\}\left(1 - \left(E(\mathbf{x}) - \sum_{C_i \in \mathcal{C}} E'_i(\mathbf{x}_{C_i})\right)\right) \quad (11)$$

where $\mathcal{E}(\mathbf{x}_{C_i})$ denotes the set of all valid energy functions of variable set $\mathbf{x}_{C_i}$. (Because evidence variables are fixed constants in inference, for simplicity, we omit explicit mention of the evidence $\mathbf{x}_E$, and the subscript $H$ in the energy term $E(\cdot)$ above and in other relevant terms in the following derivation. In should be clear that, in situations where such subscripts are omitted, $\mathbf{x}$ and related symbols denote only the hidden variables.) The solution to this problem leads to Theorem 3, which we restate here (with evidence symbol and hidden variable subscripts omitted).

**Theorem (GMF)**: *For a general undirected probability model* $p(\mathbf{x})$ *and a clustering* $\mathcal{C} : \{\mathbf{x}_{C_i}\}_{i=1}^{I}$, *if all the potential functions that cross cluster borders are cluster-factorizable, then the generalized mean*

---

[1] Note that (9) is very similar to the Jensen bound on log likelihood: $\ln p(\mathbf{x}_E) \geq \int d\mathbf{x}_H q(\mathbf{x}_H) \ln \frac{q(\mathbf{x}_H)}{p(\mathbf{x}_H, \mathbf{x}_E)}$, and has the same maximizer, but it is more general in that it can be further upgraded to higher order bounds as discussed in the discussion session.

*field approximation to $p(\mathbf{x})$ with respect to clustering $\mathcal{C}$ is a product of cluster marginals* $q^{GMF}(\mathbf{x}) = \prod_{C_i \in \mathcal{C}} q_i^{GMF}(\mathbf{x}_{C_i})$ *satisfying the following generalized mean field equations:*

$$q_i^{GMF}(\mathbf{x}_{C_i}) = p(\mathbf{x}_{C_i}|\mathcal{F}_i), \quad \forall i.$$

To prove Theorem 3 we need to use the calculus of variations to solve the optimization defined by Eq. (11). For convenience, we distinguish two subsets of nodes in a cluster $i$, the interior nodes and the border nodes, i.e., letting $\mathbf{z}_{C_i}$ denote the (hidden) nodes in cluster $C_i$, we have $\mathbf{z}_{C_i} = \{\mathbf{x}_{C_i}, \mathbf{y}_{C_i}\}$ where $\mathbf{x}_{C_i} \not\subset \mathbf{x}_{B_i}$ and $\mathbf{y}_{C_i} \subset \mathbf{x}_{B_i}$.

**Proof.** From Eq. (11), to find the optimizer of:

$$\int d\mathbf{x} d\mathbf{y} \exp\{-\sum_{C_i \in \mathcal{C}} E_i'(\mathbf{x}_{C_i}, \mathbf{y}_{C_i})\}(1 - \Delta),$$

where $\Delta \equiv E - \sum_{C_i \in \mathcal{C}} E_i' + A(\boldsymbol{\theta})$, subject to the constraints that each $E_i'$ defines a valid marginal distribution $q_i(\mathbf{x}_{C_i}, \mathbf{y}_{C_i})$ of all hidden variables in cluster $i$, we solve the Euler equations for a variational extremum, defined over Lagrangians $f(E_i', \mathbf{z}_{C_i}) = \int d\mathbf{z}_{[\cdot \setminus i]} [\exp\{-\sum_i E_i'\}(1 - \Delta) - \sum_i \lambda_i \exp\{-E_i'\}]$ (where $\mathbf{z}_{[\cdot \setminus i]}$ refers to all hidden variables excluding those from cluster $i$):

$$\frac{\partial f}{\partial E_i'} - \frac{d}{d\mathbf{z}_{C_i}}\left(\frac{\partial f}{\partial \dot{E}'_i}\right) = 0 \quad \forall i. \quad (12)$$

Since $f$ does not depend on $\dot{E}'_i$ ($= \frac{dE_i'}{d\mathbf{z}_{C_i}}$), we have:

$$\int d\mathbf{z}_{[\cdot \setminus i]} \prod_{j \neq i} \exp\{-E_j'\}(E - \sum_i E_i') - \lambda_i = 0$$

$$\Rightarrow$$

$$\begin{aligned}
E_i' &= \int d\mathbf{z}_{[\cdot \setminus i]} \prod_{j \neq i} \exp\{-E_j'\}(E - \sum_{j \neq i} E_j') - \lambda_i \\
&= C - \sum_{D_\alpha \subseteq C_i} \theta_\alpha \phi_\alpha(\mathbf{x}_{D_\alpha}) \\
&\quad - \sum_{D_\beta \in \mathcal{B}_i} \theta_\beta \langle \phi_\beta(\mathbf{y}_{C_i \cap D_\beta}, \{\mathbf{y}_{C_j \cap D_\beta}\}_{C_j \in \mathcal{N}_{i\beta}}) \rangle_{q_{\mathcal{N}_{i\beta}}},
\end{aligned}$$

where $q_j = \exp\{-E_j'(\mathbf{x}_{C_j}, \mathbf{y}_{C_j})\}$ is the local marginal of cluster $j$; $q_{\mathcal{N}_{i\beta}} = \prod_{j \in \mathcal{N}_{i\beta}} q_j$ is the marginal over cluster set $\mathcal{N}_{i\beta}$, which are all the clusters neighboring cluster $i$ that intersect with clique $\beta$.

When the potential functions at the cluster boundaries factorize with respect to the clustering, we have:

$$E_i' = C - \sum_{D_\alpha \subseteq C_i} \theta_\alpha \phi_\alpha(\mathbf{x}_{D_\alpha})$$

$$- \sum_{D_\beta \in \mathcal{B}_i} \theta_\beta F_\beta(\phi_{\beta_i}(\mathbf{y}_{C_i \cap D_\beta}), \{(\phi_{\beta_j}(\{\mathbf{y}_{C_j \cap D_\beta}\}))_{q_j}\}_{C_j \in \mathcal{N}_{i\beta}})$$

$$\begin{aligned}
\text{So,} \quad q_i(\mathbf{x}_i, \mathbf{y}_i) &= \exp\{-E_i'\} \\
&= p(\mathbf{x}_{C_i}, \mathbf{y}_{C_i} | \{(\phi_{\beta_j}(\mathbf{y}_{C_j \cap D_\beta}))_{q_j}\}_{C_j \in \mathcal{N}_{i\beta}, D_\beta \in \mathcal{B}_i}) \\
&= p(\mathbf{z}_{C_i} | \mathcal{F}_i), \quad \forall i. \quad (13)
\end{aligned}$$

The explicit presence of evidence $\mathbf{x}_E = \{\mathbf{x}_{E,C_i}\}_{i=1}^I$ merely changes Eq. (13) to $q_i(\mathbf{z}_{C_i}) \propto p(\mathbf{z}_{C_i}, \mathbf{x}_{E,C_i}|\mathcal{F}_i)$. After normalization, it leads to

$$q_i(\mathbf{z}_{C_i}) = p(\mathbf{z}_{C_i}|\mathbf{x}_{E,C_i}, \mathcal{F}_i).$$

∎